\pdfoutput=1

\documentclass[11pt]{article}

\usepackage{emnlp2021}

\usepackage{times}
\usepackage{latexsym}
\usepackage{enumitem}
\usepackage{amsmath,graphicx}
\usepackage{multirow}
\usepackage{booktabs}
\usepackage{color}
\usepackage{microtype}      

\usepackage[T1]{fontenc}

\usepackage[utf8]{inputenc}

\usepackage{microtype}

\title{FastCorrect 2: Fast Error Correction on Multiple Candidates for \\ Automatic Speech Recognition}

\author{
  Yichong Leng$^1$\thanks{\ \ This work was conducted at Microsoft. Corresponding author: Xu Tan, xuta@microsoft.com}, Xu Tan$^2$, Rui Wang$^2$, Linchen Zhu$^3$, Jin Xu$^4$, Wenjie Liu$^3$\\
  \textbf{Linquan Liu$^3$, Xiang-Yang Li$^1$, Tao Qin$^2$, Edward Lin$^3$, Tie-Yan Liu$^2$}\\
  $^1$University of Science and Technology of China, $^2$Microsoft Research Asia\\
  $^3$Microsoft Azure Speech,$^4$Tsinghua University\\
  \texttt{$^1$lyc123go@mail.ustc.edu.cn,xiangyangli@ustc.edu.cn}\\
  \texttt{$^2$\{xuta,ruiwa,taoqin,tyliu\}@microsoft.com}\\
  \texttt{$^3$\{linczhu,liwenjie,linqul,edlin\}@microsoft.com}\\
  \texttt{$^4$j-xu18@mails.tsinghua.edu.cn}\\
  \url{https://github.com/microsoft/NeuralSpeech}
}

\begin{document}
\maketitle
\begin{abstract}
Error correction is widely used in automatic speech recognition (ASR) to post-process the generated sentence, and can further reduce the word error rate (WER). Although multiple candidates are generated by an ASR system through beam search, current error correction approaches can only correct one sentence at a time, failing to leverage the voting effect\footnote{See the second paragraph in Section \ref{sec:intro}.} from multiple candidates to better detect and correct error tokens. In this work, we propose FastCorrect 2, an error correction model that takes multiple ASR candidates as input for better correction accuracy. FastCorrect 2 adopts non-autoregressive generation for fast inference, which consists of an encoder that processes multiple source sentences and a decoder that generates the target sentence in parallel from the adjusted source sentence, where the adjustment is based on the predicted duration of each source token. However, there are some issues when handling multiple source sentences. First, it is non-trivial to leverage the voting effect from multiple source sentences since they usually vary in length. Thus, we propose a novel alignment algorithm to maximize the degree of token alignment among multiple sentences in terms of token and pronunciation similarity. Second, the decoder can only take one adjusted source sentence as input, while there are multiple source sentences. Thus, we develop a candidate predictor to detect the most suitable candidate for the decoder. Experiments on our inhouse dataset and AISHELL-1 show that FastCorrect 2 can further reduce the WER over the previous correction model with single candidate by 3.2\% and 2.6\%, demonstrating the effectiveness of leveraging multiple candidates in ASR error correction. FastCorrect 2 achieves better performance than the cascaded re-scoring and correction pipeline and can serve as a unified post-processing module for ASR.
\end{abstract}

\section{Introduction}
\label{sec:intro}

Error correction has been applied in automatic speech recognition (ASR), which post-processes the outputs of the ASR system to achieve lower word error rate (WER)~\cite{ringger2001error,cucu2013stat,d2016automatic,tanaka2018neural}. Taking the recognized sentence from the ASR system as source and the ground-truth sentence as target, ASR correction can be formulated as a sequence-to-sequence problem and modeled with autoregressive \cite{mani2020asr,liao2020improving} or non-autoregressive \cite{leng2021fastcorrect} generation.

The key challenge in ASR error correction is to detect and correct the error tokens. Current approaches only correct one sentence at a time, which might be sub-optimal since the correction model can only \textit{guess} the error token based on the context information of a single sentence. Considering that beam search is commonly used in ASR, multiple candidates are usually generated and available for error correction. We argue that multiple candidates contain the \textit{voting effect}, which refers to that the tokens from multiple sentences can verify the correctness with each other. For example, if the beam search candidates with 3 sentences are ``I have cat'', ``I have hat'', ``I have bat'', then the first two tokens are likely to be correct since they are the same among all beam candidates. The inconsistency on the last token shows that: 1) this token may need correction, and 2) the pronunciation of the ground-truth token may 
end with "æt". This voting effect can be utilized to boost the ASR correction by helping the model detect error token and giving some clues about the pronunciation of ground-truth token.

In this paper, we propose FastCorrect 2, an error correction model that takes multiple ASR candidates as input to leverage this voting effect for better correction accuracy. In order to satisfy the latency constraint for industrial deployment, we leverage non-autoregressive (NAR) generation~\cite{gu2019levenshtein,lee2018deterministic,guo2019non,guo2020fine,leng2021fastcorrect} for fast inference. FastCorrect 2 consists of an encoder that processes multiple source sentences, a duration predictor to predict the number of target tokens corresponding to each source token, and a decoder that generates the target sentence in parallel from the adjusted source sentence, where the adjustment is based on the predicted duration of each source token. We describe the challenges when supporting multiple candidates and introduce our corresponding designs to address these challenges as follows.
\begin{itemize}[leftmargin=*]
\item Since the lengths of multiple candidates usually vary and the tokens from different sentences are not aligned by position, it is non-trivial to align these candidates by tokens in order to leverage the voting effect. If we simply use left or right padding to ensure the same length for alignment, the information of each position in different candidates is not aligned, and thus the voting effect does not exist. For example, if a sentence contains an extra word at the beginning while other sentences do not, naive left padding will cause severe dislocation and every position will contains misaligned tokens, making it hard for the model to detect the error tokens. To take the advantage of the voting effect, we propose a novel alignment algorithm based on token matching score and pronunciation similarity score, which can ensure the tokens on the same position are matched as much as possible, and ensure the pronunciations of tokens on the same position as similar as possible if tokens are not matched. The aligned candidates are mapped to embedding, concatenated by position and fed into the encoder of the correction model.

\item There are multiple candidates as source sentences, while the decoder can only take one adjusted source sentence as input\footnote{Since the predicted duration might be different in different candidates during inference, it is intractable to feed all adjusted candidates with different length into the decoder.}. Thus, how to choose the appropriate source sentence to adjust and take as input to the decoder is necessary. Therefore, we design a candidate predictor to decide the most appropriate source sentence. Specifically, we choose the candidate that can yield the smallest loss (i.e., the easiest candidate to correct) in the correction model.
\end{itemize}

We conduct experiments on internal ASR datasets and public AISHELL-1 datasets to verify the effectiveness of FastCorrect 2. Experiment results show that our method 1) achieves better correction accuracy over previous non-autoregressive correction model on single sentence, and 2) achieves better performance than the conventional cascaded re-scoring and correction pipeline and can serve as a unified post-processing module for ASR.

The contributions of FastCorrect 2 are summarized as follows:
\begin{itemize}[leftmargin=*]
\item We introduce multiple candidates generated by ASR beam search to help the error correction model, which can better detect the error tokens and determine the pronunciations of the ground-truth tokens, and thus can significantly improve the correction accuracy. Moreover, by utilizing all the beam search results, additional re-scoring procedure is not needed, which is time-efficient.
\item We develop a novel alignment algorithm based on token matching score and pronunciation similarity score to align the beam search candidates with variant lengths. We modify the architecture of the encoder and introduce a candidate predictor into non-autoregressive correction model to handle multiple candidates.
\end{itemize}

\section{Background}
\label{gen_inst}
In this section, we briefly review the two post-processing methods for ASR: error correction and re-scoring.

\subsection{Error Correction}
In ASR, error correction has been widely used as a post-processing method to improve the quality of recognized text \cite{tanaka2018neural,anantaram2018repairing,shivakumar2019learning}. Considering that the input and output domain of ASR correction are both text, utilizing sequence-to-sequence technologies becomes a popular direction. \citet{cucu2013stat} leveraged statistic machine translation and \citet{d2016automatic} used phrase-based machine translation system for ASR correction. With the development of attention mechanism, \citet{mani2020asr} trained an autoregressive correction model with Transformer \cite{vaswani2017attention} architecture. \citet{liao2020improving} further incorporated pre-training method into ASR correction. 
In order to meet the latency requirements for industrial applications of ASR, \citet{leng2021fastcorrect} proposed a non-autoregressive model to accelerate the correction without performance deterioration.

In general, current correction methods can only correct errors of a single sentence by leveraging the context information of this single sentence. In this work, to further reduce the word error rate of ASR while satisfying the latency requirements for industrial deployment, we propose FastCorrect 2 by taking advantage of the voting effect in the multiple candidates of ASR, which can help better detect the error tokens and thus benefit error correction.

\subsection{Re-scoring}
Re-scoring is used as a post-processing technology for ASR to select the best candidate from the multiple candidates generated by beam search with an external neural language model. The selection criteria is the ranking score, which is a weighted linear combination of the acoustic model score from ASR system and the external language model score.

Given that the correction model focuses on improving the quality of one candidate and the re-scoring involves choosing the best candidate, these two post-processing methods are not mutually exclusive and can be combined sequentially to further improve the ASR quality. Since FastCorrect 2 takes multiple candidates as input, it is not necessary to add an additional re-scoring procedure, showing that FastCorrect 2 can serve as a time-efficient unified post-processing method for ASR.

\begin{figure*}
  \centering
  \includegraphics[width=.9\textwidth]{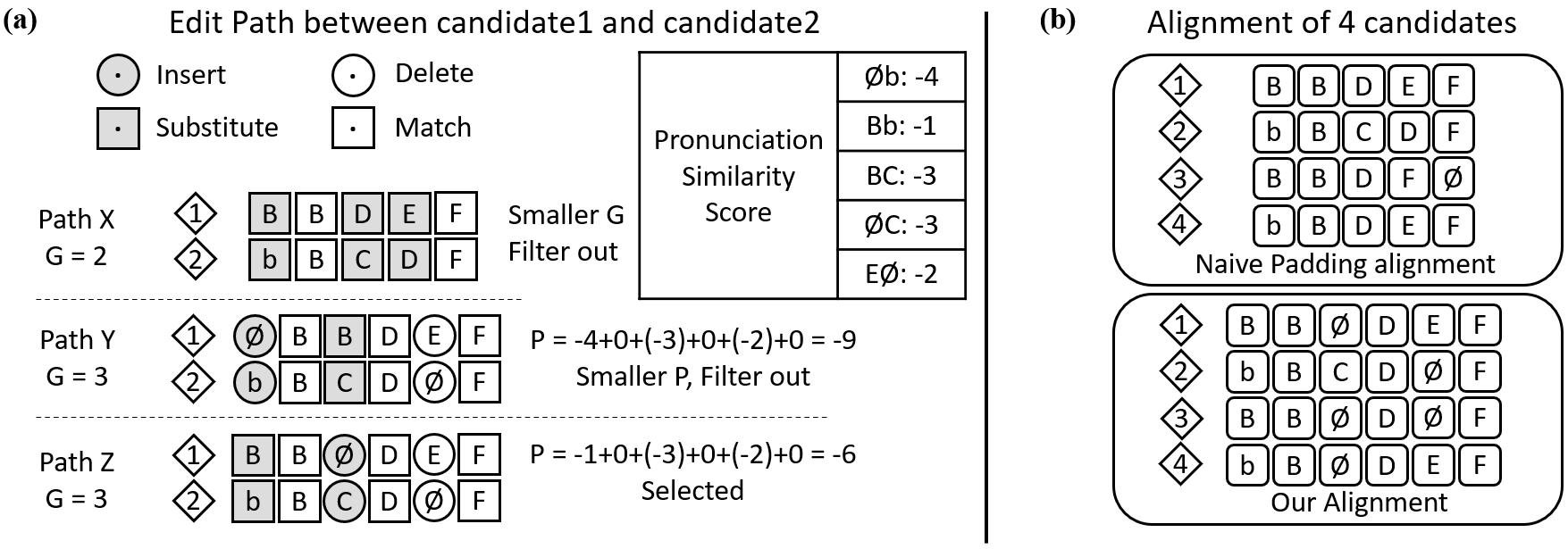}
  \caption{The proposed alignment method to align multiple candidates. Sub-figure (a) shows the detailed procedure when aligning the first candidate ``B B D E F'' with the second candidate ``b B C D F''. Each character stands for a token, and we denote ``b'' as a different token from ``B'' but with similar pronunciation. We enumerate all edit paths between two candidates with the smallest edit distance, then we calculate the token matching score and pronunciation similarity score of each path, selecting the path with highest token matching score first and then highest pronunciation similarity score. Taking path Z as an example, the token matching score G is 3 because it contains 3 positions with the same token (i.e., ``B D F'') and the pronunciation similarity score P of a path is the sum of all pronunciation similarity score of each position, which is shown in pronunciation similarity score table. Sub-figure (b) shows the comparison between the proposed alignment method and the naive alignment (padding to right), which shows that the proposed alignment can keep the token and pronunciation consistent on each position (in contrast to the 4th position in naive padding alignment which consists of 3 different tokens ``E D F'').}
  \label{fig:align}
\end{figure*}

\section{FastCorrect 2}
\label{headings}
In this section, we introduce FastCorrect 2 in detail. In order to leverage the voting effect in multiple candidates, we first propose an alignment algorithm to align the length-variant candidates into the same length while maximizing the token and pronunciation similarity on each alignment position. Then we introduce the detailed architecture of our model, consisting of an encoder with a Pre-Net to handle the multiple candidates input, a duration predictor to predict the number of target tokens corresponding to each source token in each candidate, and a candidate predictor to find out the easiest candidate to correct, which is then adjusted according to the predicted duration and used as the decoder input for parallel generation. We introduce the alignment algorithm and the correction model in the following subsections.

\subsection{Alignment for Multiple Candidates}
In general, when aligning $n$ candidates $\{b_1, b_2, ..., b_n\}$, we first randomly select a candidate ${b_1}$ as the anchor candidate, and then align the remaining candidates with it, resulting in $n-1$ alignments $\{a_{12}, a_{13}, ..., a_{1n}\}$. The final alignment on all candidates can be obtained by merging $n-1$ alignments. The core of the proposed alignment algorithm includes how to align two candidates and how to merge alignments, which are discussed in detail as follows.

\paragraph{Alignment for Two Candidates} As shown in Figure \ref{fig:align}a, we can calculate the edit distance\footnote{The edit distance of two sequence is the minimum number of operations needed to edit one sequence to match the other sequence. The operation includes insertion, deletion and substitution.} between two candidates and enumerate all possible edit paths (i.e., sequences of operations including insertion, deletion and substitution) with the minimum edit distance. For insertion and deletion operation, a special token ``Ø'' is added indicating an empty token, which can help derive the alignment from the edit path (e.g., path X/Y/Z in Figure \ref{fig:align}a).

We select the final alignment by calculating the token matching score and the pronunciation similarity score of each path. Specifically, the token matching score is the number of positions whose tokens are all the same. The pronunciation similarity score of a path is the sum of pronunciation similarity score of all the token pairs, which is defined as the negative value of edit distance between their phoneme sequence.
The paths with highest token matching score are first selected, and among which the path with the highest pronunciation similarity score is chosen. The token matching score has higher priority than the pronunciation similarity score when choosing the paths because the same token has the same pronunciation.

\paragraph{Merging Alignments} For the $n-1$ alignments $\{a_{12}, a_{13}, ..., a_{1n}\}$ obtained from the above step, we can merge them to get a final alignment for all the candidates. Considering that every token in anchor candidate $b_1$ appears in the $n-1$ alignments, we can merge the alignments by regarding the tokens in $b_1$ as anchor points. If the anchor candidate $b_1$ contains an empty token, which is introduced in the above step to help for the alignment, this empty token will also be regarded as an anchor token.

Figure \ref{fig:align}b illustrates the advantages of the proposed alignment method compared to the naive alignment method (padding to right). Our alignment method can 1) align the same tokens (``B'', ``D'' and ``F'') at the same position, 2) isolate the additional token occurred only in one candidate (``C''), and 3) keep the pronunciation similarity of the tokens on the same position as high as possible, which can help the correction model to detect the error token and infer the ground-truth pronunciation of the token.

\subsection{Proposed Model Architecture}
\begin{figure}
  \centering
  \includegraphics[width=.48\textwidth]{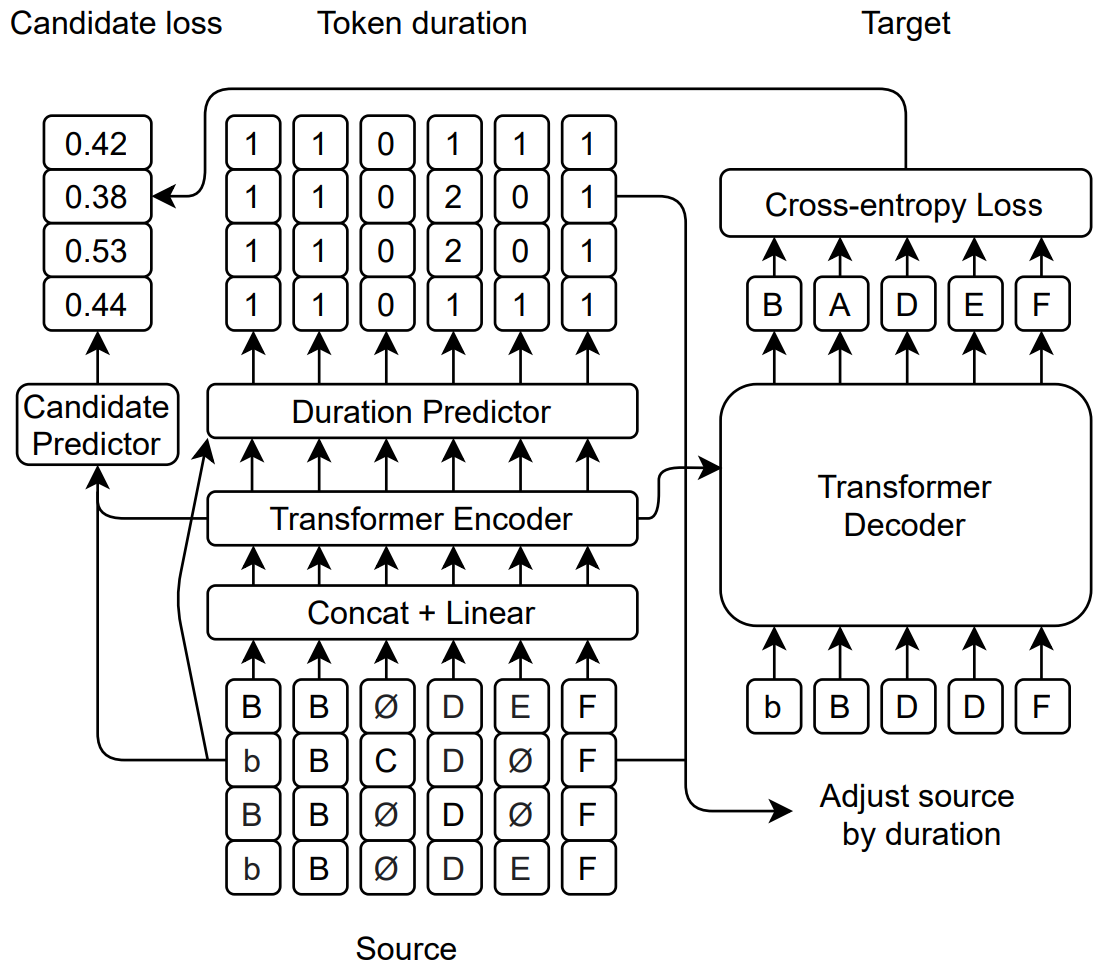}
  \caption{Model architecture. The aligned beam search results are concatenated along position, reshaped by a linear layer and then fed into encoder. The encoder output is concatenated with the original token embedding and fed into predictor to predict the duration of each source token (by duration predictor) and the loss of candidate (by candidate predictor). The source token is adjusted according to the duration predictor and then fed into decoder. Finally, the loss of the decoder is used as the label of candidate predictor. Please note that the encoder input is the merged information of all candidates while the duration and candidate predictor is applied on every candidate, respectively.}
  \label{fig:nat_model}
\end{figure}

In this section, we provide a detailed description of the architecture of FastCorrect 2, as shown in Figure \ref{fig:nat_model}. The backbone of FastCorrect 2 is based on FastCorrect~\cite{leng2021fastcorrect}, which is fast and accurate for ASR error correction. FastCorrect utilizes a duration predictor to predict the duration of each source token (i.e., the number of target tokens corresponding to a certain source token), based on which the source toke is adjusted and fed into decoder for parallel generation. We follow the algorithm in FastCorrect to extract the duration of each source token \cite{leng2021fastcorrect}, and set the duration of empty token ``Ø'' as 0. 
To enable the correction model to benefit from the multiple candidates, we introduce or modify some modules in FastCorrect architecture as follows.

\paragraph{Encoder} Since the aligned candidates have high token matching score and pronunciation similarity score on each position, the encoder should handle the tokens on the same position together to make fully use of the voting effect. So, we add a Pre-Net in the Transformer encoder, which concatenates the token embedding of all candidates on each position and uses a linear layer to reshape the concatenated feature to the encoder hidden size.

\paragraph{Duration Predictor} Although the candidates has been aligned to the same length, the number of target tokens corresponding to each source token in different candidates still varies\footnote{As shown in Figure \ref{fig:nat_model}, the duration of token ``D'' in the first candidate is 1 but that in the second candidate is 2.}. Therefore, it is sub-optimal to directly use duration predictor to predict the token duration of different candidates based on the encoder output containing merged information of all candidates. To ensure the duration predictor to be more discriminative, we concatenate the encoder output and the original encoder input (different for each candidate) on each position and take them as input to the duration predictor.

\paragraph{Candidate Predictor} We introduce a candidate predictor to enable model to choose the easiest candidate for correction. The candidate predictor is trained to predict the loss of correction model on each candidate, whose label is the cross-entropy loss of decoder output. The input of candidate predictor is the same as duration predictor, which is a concatenation of encoder output and corresponding candidate embedding.

\paragraph{Decoder} We use the same Transformer decoder architecture as FastCorrect, which takes the adjusted source token of one candidate as input and generates the corrected sentence in parallel.

\section{Experiment Settings}
In this section, we show the datasets, ASR models, settings of FastCorrect 2, as well as the baseline systems used in experiments.

\subsection{Dataset}
We evaluate FastCorrect 2 on two datasets,  the public dataset AISHELL-1 \cite{bu2017aishell}\footnote{https://openslr.org/33} and an inhouse dataset. The AISHELL-1 dataset is a Mandarin speech corpus with 178 hours training data, 10 hours validation data and 5 hours test data. Our inhouse dataset is a large industrial ASR dataset consisting of 75K hours Mandarin speech, the size of validation and test set are both 200 hours. Once the ASR models are trained on these datasets, the models are used to transcribe the training corpus, where the transcribed sentences and the ground-truth sentences construct the training corpus for ASR correction model. The beam size used in transcribing is set to 4, which means that FastCorrect 2 takes the aligned 4 beam candidates as model input.

From the previous work \cite{leng2021fastcorrect}, correction model can hardly achieve good correction ability on a small-scale dataset (e.g., AISHELL-1) without pretraining on pseudo data. We crawl 400M unpaired text from the internet and randomly add noise (insertion, deletion or substitution with a homophone dictionary) to the text to construct the pseudo dataset for AISHELL-1 pretraining. The ratio of noise and the probability distribution of noise type (insertion, deletion and substitution) are determined by the word error rate (WER) and statistics of the training corpus for correction, which is constructed by using an ASR model to transcribe the AISHELL-1 training set as mentioned in the above paragraph. 

We use SentencePiece \cite{kudo2018sentencepiece} to learn subword and apply to all the text above. The dictionary size is set to 40K.

\subsection{ASR Model}
\label{subsec:asr_model}
We use the ESPnet \cite{watanabe2018espnet} toolkit to train an ASR model on AISHELL-1 dataset. In order to verify FastCorrect 2 in a competitive setting, we utilize several advanced techniques to train a strong ASR model, including Conformer architecture \cite{gulati2020conformer}, SpecAugment \cite{park2019specaugment}, and speed perturbation data augmentation. The language model used in inference is a strong Transformer-based model trained on the crawled 400M dataset. The ASR model achieves a state-of-the-art character error rate (CER) of 4.03 and 4.31 on the validation and test set of AISHELL-1.

For inhouse dataset, we train an industrial ASR model with highly competitive accuracy, where the acoustic model is a latency-controlled BLSTM \cite{zhang2016highway} with 6 layers and 1024 hidden units in each layer. We use the inhouse dataset to show that FastCorrect 2 can 1) meet the industrial requirements, 2) still be effective when the training corpus is large.

\subsection{FastCorrect 2 Model}
In FastCorrect 2, the layer number and hidden size of Transformer is 6, 512, respectively. The duration predictor consists of 5 layers of 1D convolutional network with ReLU activation and 2 linear layers to output a scalar, all of which has a hidden size of 512. Each convolutional layer is followed by layer normalization \cite{ba2016layer} and dropout. The kernel size of the convolutional network is 3. The candidate predictor additionally contains a global mean pooling layer to predict the candidate loss.

We train all correction models on 8 NVIDIA V100 GPUs, with a batch size of 6000 tokens. We use standard training hyper-parameters of Transformer in Fairseq \cite{ott2019fairseq}. To simulate the industrial scenario, we test the inference speed of correction models in three conditions: 1) NVIDIA P40 GPU, 2) 1-core CPU, and 3) 4-core CPU\footnote{Intel(R) Xeon(R) CPU E5-2690 v4 @ 2.60GHz.}. The test batch size is set to 1 to simulate the online serving condition.

\subsection{Baseline Systems}
We compare FastCorrect 2 with several correction models based on both autoregressive architecture and non-autoregressive architecture. Since FastCorrect 2 takes advantage of beam search candidates, we also compare with the cascaded pipeline of correction and re-scoring.
\paragraph{FastCorrect} We train the FastCorrect baseline following the setting in \citet{leng2021fastcorrect}, the layer number and hidden size are set to be the same as FastCorrect 2.
\paragraph{Autoregressive Correction Model} We train an autoregressive correction model, whose architecture is the same as FastCorrect 2 except that AR model has no candidate predictor and duration predictor.
\paragraph{Cascaded Pipeline} There are two versions of cascade pipeline: 1) R+FC, where we re-score multiple candidates first and then perform correction on one candidate. 2) FC+R, where we perform correction on all candidates first and then use re-scoring to choose the best candidate. 

Roughly speaking, the latency of FC+R is $n$ times larger than R+FC, where $n$ is the beam size, since the correction in FC+R is applied to all $n$ candidates, while in R+FC, correction is only applied to 1 candidate. Meanwhile, the accuracy of C+R is likely to be higher because the re-scoring is performed on candidates with higher quality. The language model used in re-scoring is the same as that in section \ref{subsec:asr_model}, which is a 12-layer Transformer encoder with a hidden size of 512.

\begin{table*}[t]
\caption{The correction accuracy and inference latency of different correction models. We report the word error rate (WER), word error rate reduction (WERR) and latency of the autoregressive correction moddel (AR Correct), FastCorrect and FastCorrect 2, as well as the two versions of the cascaded pipeline. ``R + FC'' stands for re-scoring firstly and then correcting by FastCorrect. ``FC + R'' stands for performing correction by FastCorrect firstly and then re-scoring, which is slower than ``R + FC''.}
\label{tab:main_result}
\vspace{-2mm}
\begin{center}
\begin{tabular}{l|c|c|c|c|ccc}

\toprule
\multirow{2}{*}{AISHELL-1} & \multicolumn{2}{c|}{Test Set}  & \multicolumn{2}{c|}{Dev Set}  & \multicolumn{3}{c}{Latency (ms/sent) on Test Set}
\tabularnewline\cmidrule{2-8}
\multicolumn{1}{c|}{} & \multicolumn{1}{c|}{WER}   & \multicolumn{1}{c|}{WERR}  & \multicolumn{1}{c|}{WER}   & \multicolumn{1}{c|}{WERR}    & GPU       & CPU*4       & CPU   \\\midrule
No correction & 4.31 & -   & 4.03 & - & -  &  -   &  -    \\
AR Correct  & 3.85 & 10.67 & 3.61 & 10.42 & 149.5 \small{(1$\times$)}     &   248.9 \small{(1$\times$)} &  531.3 \small{(1$\times$)} \\
FastCorrect  & 3.95 & 8.35 & 3.69 & 8.44 & 21.2 \small{(7.1$\times$)}    &   40.8 \small{(6.1$\times$)}   &  82.3 \small{(6.5$\times$)} \\\midrule
R + FC  & 3.93 & 8.82 & 3.65 & 9.43 & 35.7 \small{(4.2$\times$)}     &   63.0 \small{(4.0$\times$)} &  124.0 \small{(4.3$\times$)}  \\
FC + R  & \textbf{3.83} & \textbf{11.14} & 3.56 & 11.67 & 99.3 \small{(1.5$\times$)}     &   185.4 \small{(1.3$\times$)} &  370.9 \small{(1.4$\times$)} \\\midrule
FastCorrect 2  & 3.84 & 10.90 & \textbf{3.50} & \textbf{13.15} & \textbf{30.1 \small{(5.0$\times$)}}    &  \textbf{ 50.0 \small{(5.0$\times$)}} & \textbf{ 106.9 \small{(5.0$\times$)}} \\
\midrule

\midrule
\multirow{2}{*}{Internal Dataset} & \multicolumn{2}{c|}{Test Set}  & \multicolumn{2}{c|}{Dev Set}  & \multicolumn{3}{c}{Latency (ms/sent) on Test Set}
\tabularnewline\cline{2-8}
\multicolumn{1}{c|}{} & \multicolumn{1}{c|}{WER}   & \multicolumn{1}{c|}{WERR}  & \multicolumn{1}{c|}{WER}   & \multicolumn{1}{c|}{WERR}    & GPU       & CPU*4       & CPU   \\\midrule
No correction & 11.17 & -   & 11.24 & - & -  &  -   &  -    \\
AR model  & 10.22 & 8.50 & 10.31 & 8.27 & 191.5 \small{(1$\times$)}     &   336  \small{(1$\times$)}  &  657.7  \small{(1$\times$)}  \\
FastCorrect  & 10.27 & 8.06 & 10.35 & 7.92  & 21.5  \small{(8.9$\times$)}   &   42.4  \small{(7.9$\times$)}  &  88.6  \small{(7.4$\times$)}  \\\midrule
R + FC  & 10.18 & 8.86 & 10.26 & 8.72  & 36.2  \small{(5.3$\times$)}   &   65.2  \small{(5.2$\times$)}  &  132.4  \small{(5.0$\times$)}  \\
FC + R  & 10.14 & 9.22 & 10.23 & 8.99 &  100.7  \small{(1.9$\times$)}   &   192.4  \small{(1.7$\times$)}  &  398.2  \small{(1.7$\times$)}  \\\midrule
FastCorrect 2  & \textbf{9.91} & \textbf{11.28} & \textbf{9.99} & \textbf{11.12} &  \textbf{32.4 \small{(5.9$\times$)}}   & \textbf{  57.1 \small{(5.9$\times$)} }   &  \textbf{118.7  \small{(5.5$\times$)}  }\\

\bottomrule

\end{tabular}
\end{center}
\vspace{-3mm}
\end{table*}

\begin{table}[t]
\caption{Ablation study of alignment algorithm.}
\label{tab:aba_align}
\vspace{-2mm}
\begin{center}
\begin{tabular}{l|c|c}
\toprule
Internal Dataset & WER & WERR\\\midrule
FastCorrect 2 & \textbf{9.91} &   \textbf{11.28}  \\
\ \ - Pronunciation score & 9.94 &   11.01  \\
\ \ - Token matching score & 9.97 &   10.74  \\
Naive padding & 10.04 &   10.11  \\
\bottomrule
\end{tabular}
\end{center}
\vspace{-5mm}
\end{table}

\begin{table}[t]
\caption{The effectiveness of our alignment algorithm on autoregressive correction model. The last row is the AR model with our alignment algorithm. }
\label{tab:at_result}
\vspace{-2mm}
\begin{center}
\begin{tabular}{l|c|c}
\toprule
\multirow{2}{*}{Model} & \multicolumn{1}{c|}{AISHELL-1} &  Internal 
\tabularnewline
\multicolumn{1}{c|}{} & \multicolumn{1}{c|}{Dataset}    & Dataset   \\\midrule
No correction &   4.31  & 11.17 \\ \midrule
AR Correct & 3.85  &  10.22  \\\midrule
Re-scoring + AR &  3.84 &   10.10  \\ 
AR + re-scoring  & \textbf{3.73} &   9.99 \\\midrule
AR + alignment  &  3.75 & \textbf{9.79} \\
\bottomrule
\end{tabular}
\end{center}
\vspace{-5mm}
\end{table}

\begin{table}[t]
\caption{Ablation study of candidate predictor.}
\label{tab:aba_predictor}
\vspace{-2mm}
\begin{center}
\begin{tabular}{l|c|c}
\toprule
\multirow{2}{*}{Choosing beam by} & \multicolumn{1}{c|}{AISHELL-1}   & Internal
\tabularnewline
\multicolumn{1}{c|}{} & \multicolumn{1}{c|}{Dataset}    & Dataset   \\\midrule
No correction &  4.31  &  11.17 \\ \midrule
Candidate predictor &  \textbf{3.84}  &  \textbf{9.91} \\
WER predictor &   3.88 &  10.01  \\ 
First beam  & 3.89 &  10.08   \\
Random  & 3.99 & 10.09  \\
\bottomrule
\end{tabular}
\end{center}
\vspace{-5mm}
\end{table}

\section{Results}
In this section, we first report the accuracy and latency of FastCorrect 2 for ASR error correction, and then perform ablation study to verify the effectiveness of each module in FastCorrect 2.

\subsection{Accuracy and Latency}
In Table \ref{tab:main_result}, we report the correction accuracy and inference latency of different correction models, from which we have the following observations:

First, compared with the FastCorrect baseline, FastCorrect 2 can improve the correction accuracy by 2.55\% and 3.22\% in terms of WER reduction on AISHELL-1 and internal dataset, respectively, which shows the effectiveness of utilizing multiple candidates information. Moreover, FastCorrect 2 is 5 times faster than the autoregressive model, indicating the inference efficiency of FastCorrect 2.

Second, FastCorrect 2 has a better performance than the fast cascaded pipeline (i.e., ``Re-score + FC''). Compared with the slow cascaded pipeline (i.e., ``FC + Re-score''), FastCorrect 2 achieves better accuracy on internal dataset and comparable accuracy on AISHELL-1, however, the latency of FastCorrect 2 is 3 times faster than the slow cascaded pipeline. The results suggest that FastCorrect 2 can unify the two post-processing methods of ASR, in a faster and better way.

\subsection{Ablation Study}
In this section, we conduct ablation study to verify the effectiveness of the alignment algorithm and the candidate predictor.

\paragraph{Alignment Algorithm} We perform ablation study on our alignment algorithm by first removing pronunciation similarity score and then further removing token matching score. Once both pronunciation similarity score and token matching score are removed, our algorithm fails back to choose the path based on its operations, whose priority is set to ``Identity > Substitution > Insertion > Deletion''. Moreover, we include the naive padding (padding to right, as shown in the Figure \ref{fig:align}b) into ablation study, whose results are in Table \ref{tab:aba_align}. Due to the computational cost of pretraining, we perform the comparison without pre-training, and we only report the results on the internal dataset because pre-training is necessary to obtain reasonable results on AISHELL-1. From this table, it can be seen that FastCorrect 2 outperforms the naive padding, showing that FastCorrect 2 can exploit more information in multiple candidates and thus improve the correction accuracy. The results also show that pronunciation similarity score and token matching score are useful to ensure the correction accuracy.

We also conduct an additional experiment to verify the effectiveness of our alignment algorithm by improving the accuracy of autoregressive correction model, and show the results in Table \ref{tab:at_result}. It can be seen that the proposed method can also reduce the WER of autoregressive correction model, which is better than the combination of correction and re-scoring on internal dataset, indicating that our alignment algorithm is a general method to exploit voting effect and improve the correction accuracy. 

\paragraph{Candidate Predictor} The core function of candidate predictor in FastCorrect 2 is to find out the easiest beam candidate for decoder to correct. We compare the candidate predictor with 3 other candidate-selection methods: 1) For each beam candidate in training set, we calculate the WER between that candidate and target. Then we train a WER predictor to predict which beam candidate has the smallest WER with the target. 2) We simply choose the first beam candidate scored by the ASR model and perform correction. 3) We randomly choose a beam candidate to correct. 

The comparisons are shown in Table \ref{tab:aba_predictor}, in which the candidate predictor outperforms the baseline methods on both datasets. Using a WER predictor is better than choosing the first beam candidate but worse than the candidate predictor because the beam candidate with lowest WER sometimes is not the easiest one for decoder to correct, showing the advantage of candidate predictor which explicitly models the correction difficulty of decoder.

\subsection{Comparison with Data Augmentation}
\begin{figure}[h]
  \centering
  \includegraphics[width=.48\textwidth]{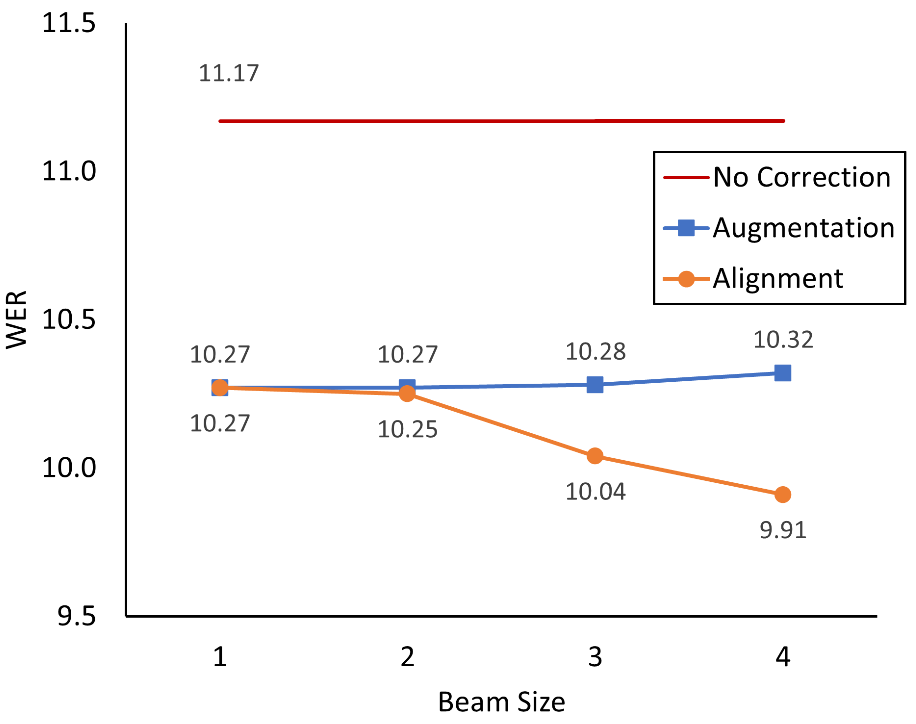}
  \caption{The WER of model with respect to candidate size on the internal dataset. The red line without mark is the WER without correction (i.e., WER=11.17). The blue line with square mark is the WER when using multiple candidates as data augmentation. The orange line with dot mask is the WER of FastCorrect 2. }
  \label{fig:wer_wrt_beam}
\end{figure}

A straightforward method to leverage the multiple candidates from beam search is using candidates as data augmentation. By pairing every candidate with its corresponding ground-truth text, we can construct a new dataset to train FastCorrect baseline. The size of new dataset is $n$ times larger, where $n$ is the beam size. We compare the accuracy of data augmentation and FastCorrect 2 in difference beam sizes, and the results are shown in Figure \ref{fig:wer_wrt_beam}. We have several observations:

First, simply using multiple candidates as data augmentation cannot yield better result comparing with only using the best multiple scored by ASR model, showing that the key component to benefit from multiple candidates is to take advantage of voting effect rather than the total data amount.
Second, FastCorrect 2 yields better result with more candidates introduced into model, which can be accounted by voting effect. When the beam size is 2, since the candidate number is too small for a \textit{vote}, FastCorrect 2 only leads to slightly accuracy improvement. In constrast, once we have large beam size, FastCorrect 2 can be aware of the clue of error token and ground-truth token pronunciation, and thus reduces the baseline WER by a large margin.

In summary, the comparison with data augmentation shows that it is necessary to use aligned beam candidates to exploit the voting effect, so as to further boost the model performance.

\subsection{Comparison with Other Methods}
We compare FastCorrect 2 with two baselines which also leverage multiple candidates in this section.

\paragraph{ROVER} ROVER \cite{fiscus1997post} is a traditional baseline\footnote{https://github.com/usnistgov/SCTK/blob/20159b580249f 1598caa35ab469bd1acdb3dd86c/doc/rover.htm} to make use of multiple candidates, which aligns multiple candidates first and then votes for the final token on each position by occurrence \cite{amith2021end}. In our experiments, we try both the default open-source alignment method and our proposed alignment method.

\paragraph{Fusion} Fusion is another method for utilizing multiple candidates \citep{liu2021asr,lohrenz2021multi}. Specifically, Fusion uses a shared encoder to extract the representation of each candidate and multiple encoder-decoder attentions to fuse the encoder output of each candidate into decoder, which has no need on alignment algorithm or candidate predictor. 

The comparison results of FastCorrect 2 with the above two baselines are shown in Table \ref{tab:other_baseline}. The results show that ROVER cannot boost the performance of ASR (even worsen the performance). However, our proposed alignment method outperforms the default alignment method of ROVER, demonstrating the effectiveness of our alignment method. When it comes to Fusion, it has an accuracy slightly superior to FastCorrect, demonstrating that it can make use of the information in multiple candidates to some extent. However, FastCorrect 2 outperforms Fusion by a larger margin, showing the effectiveness of our alignment algorithm and candidate predictor.

\begin{table}[t]
\caption{Comparison with other methods.}
\label{tab:other_baseline}
\vspace{-2mm}
\begin{center}
\begin{tabular}{l|c|c}
\toprule
\multirow{2}{*}{Model} & \multicolumn{1}{c|}{AISHELL-1} &  Internal 
\tabularnewline
\multicolumn{1}{c|}{} & \multicolumn{1}{c|}{Dataset}    & Dataset   \\\midrule
No correction &   4.31  & 11.17 \\
FastCorrect &   3.95  & 10.27 \\ \midrule
ROVER & 5.24  &  11.53  \\
\ \ \ \ + our alignment &  4.91 &   11.40  \\
Fusion  & 3.92 &   10.17 \\
FastCorrect 2  &  \textbf{3.84} & \textbf{9.91} \\
\bottomrule
\end{tabular}
\end{center}
\vspace{-5mm}
\end{table}

\section{Conclusion}
In this work, we propose FastCorrect 2 to leverage multiple candidates for ASR error correction, where the candidates are generated by the ASR model through beam search and contain voting effect to help better detect and correct error tokens. In order to leverage this voting effect from multiple candidates, we propose a novel alignment algorithm to align the multiple candidates while maximizing the token matching score and pronunciation similarity score on each alignment position. Accordingly, we make several modifications on a previous non-autoregressive error correction model to make it suitable for the input with multiple candidates, including an encoder with a Pre-Net to handle the concatenation of multiple candidates, a duration predictor to predict the token duration of each candidate, and a candidate predictor to choose the easiest candidate as the decoder input for correction. Experiment results show that FastCorrect 2 improves the correction accuracy over previous non-autoregressive correction model with single candidate, demonstrating the effectiveness of multiple candidates for error correction. Besides, FastCorrect 2 can be used as a unified post-processing module for ASR, achieving comparable or better accuracy with the cascaded post-processing pipeline (correction and re-scoring) while speeding up for 3 times.

\bibliography{anthology,custom}
\bibliographystyle{acl_natbib}

\end{document}